\documentclass[10pt,twocolumn,letterpaper]{article}

\usepackage{cvpr}              % To produce the CAMERA-READY version
\usepackage{subcaption}
\usepackage{graphicx}

\usepackage[dvipsnames]{xcolor}

\definecolor{cvprblue}{rgb}{0.21,0.49,0.74}
\usepackage[pagebackref,breaklinks,colorlinks,citecolor=cvprblue]{hyperref}
\newcommand{\tablestyle}[2]{\setlength{\tabcolsep}{#1}\renewcommand{\arraystretch}{#2}\centering\small}

\usepackage{tabulary,multirow,xspace}
\usepackage{subcaption}
\usepackage{caption}
\usepackage{float}
\usepackage{wrapfig} 
\usepackage[linesnumbered,ruled,vlined]{algorithm2e}
\usepackage{makecell}
\usepackage{pifont}
\newcommand{\cmark}{\ding{51}}%
\newcommand{\xmark}{\ding{55}}%
\usepackage{tcolorbox}
\usepackage{subcaption}
\usepackage{graphicx}
\usepackage{tcolorbox}

\usepackage[T1]{fontenc}

\definecolor{plotblue}{rgb}{0.235, 0.471, 0.847}
\definecolor{plotred}{rgb}{0.796, 0.075, 0.149}
\definecolor{plotbrown}{rgb}{0.523, 0.223, 0}
\definecolor{plotpurple}{rgb}{0.8, 0.6, 1}
\definecolor{plotsamplepurple}{rgb}{0.8, 0.6, 1}
\definecolor{plotsampleblue}{rgb}{0.6, 0.8, 1}
\definecolor{plotsamplegreen}{rgb}{0.2, 0.784, 0.2}
\definecolor{plotsamplered}{rgb}{0.6, 0, 0}
\definecolor{plotblue}{rgb}{0.235, 0.471, 0.847}
\definecolor{plotred}{rgb}{0.796, 0.075, 0.149}
\definecolor{plotpurple}{rgb}{0.8, 0.6, 1}
\definecolor{plotsamplepurple}{rgb}{0.8, 0.6, 1}
\definecolor{plotsampleblue}{rgb}{0.6, 0.8, 1}
\definecolor{plotsamplegreen}{rgb}{0.2, 0.784, 0.2}
\definecolor{plotsamplered}{rgb}{0.6, 0, 0}
\definecolor{cpurple}{HTML}{8330a0} %7030A0
\definecolor{plum}{HTML}{9c277e} %9C276A

\DeclareSymbolFont{extraup}{U}{zavm}{m}{n}
\DeclareMathSymbol{\vardiamond}{\mathalpha}{extraup}{87}

\def\paperID{} % *** Enter the Paper ID here
\def\confName{3DV\xspace}
\def\confYear{2025\xspace}

\title{OpenMaskDINO3D : Reasoning 3D Segmentation via Large Language Model}

\author{Kunshen Zhang$^1$ \vspace{0.2cm}\\
$^1$Wuhan University \vspace{0.2cm}\\
\url{https://github.com/Zhangkuns/OpenMaskDINO3D}
}

\begin{document}

\twocolumn[{
  \renewcommand\twocolumn[1][]{#1}
  \maketitle  
}]

\begin{abstract}
Although perception systems have made remarkable advancements in recent years, particularly in 2D reasoning segmentation, these systems still rely on explicit human instruction or pre-defined categories to identify target objects before executing visual recognition tasks. Such systems have matured significantly, demonstrating the ability to reason and comprehend implicit user intentions in two-dimensional contexts, producing accurate segmentation masks based on complex and implicit query text. 
However, a comparable framework and structure for 3D reasoning segmentation remain absent.
This paper introduces OpenMaskDINO3D, a LLM designed for comprehensive 3D understanding and segmentation. OpenMaskDINO3D processes point cloud data and text prompts to produce instance segmentation masks, excelling in many 3D tasks.
By introducing a SEG token and object identifier, we achieve high-precision 3D segmentation mask generation, enabling the model to directly produce accurate point cloud segmentation results from natural language instructions.
Experimental results on large-scale ScanNet datasets validate the effectiveness of our OpenMaskDINO3D across various tasks.
\end{abstract}    
\section{Introduction}
\label{sec:intro}
In recent years, large language models (LLMs) have achieved significant advancements in complex reasoning capabilities within the field of natural language processing. Building on these developments, a new class of models, multimodal large language models (MLLMs), has emerged. These models are designed to handle multiple modalities, including 2D images, thereby enhancing the ability of LLMs to interpret and understand visual inputs. However, while MLLMs perform exceptionally well in processing 2D images, extending these capabilities to complex 3D environments remains a significant challenge, which is critical for various technical domains such as robotic navigation and embodied AI agents.

Compared to the millions of data samples used for training 2D MLLMs, existing 3D scene-language datasets contain only tens of thousands of localization instances, making it difficult to support effective learning of positional markers for 3D MLLMs, especially given the exponentially greater complexity of 3D spaces compared to 2D. There is a severe shortage of 3D data. The annotation cost for 3D data is significantly higher than for 2D data, resulting in smaller dataset sizes, which makes training 3D MLLMs more challenging. Additionally, due to the inherent sparsity and unstructured nature of point clouds, as well as the severe scarcity of 3D scene-language data, directly adapted methods often yield poor results.

To address these issues, this paper attempts to introduce techniques similar to those in the 2D domain, such as using a SEG token to guide the decoder in learning to create 3D segmentation masks or adopting positional markers to achieve object localization, akin to methods used in 2D reasoning models. Through model-guided segmentation, we achieve a synergistic mechanism for object referencing and localization, thereby enhancing the scene understanding capabilities of 3D MLLMs.

Specifically, this paper designs an innovative approach that cleverly leverages the semantic understanding capabilities of large language models to guide precise object segmentation in 3D scenes. We introduce a special SEG token as a learning signal for the decoder, enabling the model to generate high-quality 3D segmentation masks based on user language descriptions. This method not only improves segmentation accuracy but also enhances the model’s ability to understand complex semantic instructions, allowing it to handle intricate spatial relationship descriptions such as “find the red chair to the left of the table.”
%%%% Summary of experiments, datasets and results.
%
\noindent{The main contributions of this work are:}

\begin{itemize}[align=right,itemindent=0em,labelsep=2pt,labelwidth=1em,leftmargin=*,itemsep=0em] 
\item We introduce OpenMaskDINO3D, a comprehensive framework for reasoning and instance segmentation within 3D scenes using language prompts. OpenMaskDINO3D processes 3D point clouds and language inputs to generate both textual outputs and detailed 3D segmentation masks. 
\item We establish a comprehensive benchmark for 3D reasoning segmentation, incorporating over one thousand 3D point cloud-instruction-mask data samples, to evaluate the model's ability to interpret intricate reasoning and world knowledge in three-dimensional environments.
\item We develop an innovative embedding-as-mask paradigm, augmented with a specialized SEG token, to unlock and enhance the 3D segmentation capabilities of multimodal large language models, achieving robust zero-shot performance and further improvement with minimal fine-tuning on 239 reasoning segmentation samples.
\end{itemize}
\section{Related Work}

\noindent{\bf Open-Vocabulary 3D Scene Understanding.} 
Open-vocabulary 3D scene understanding aims to identify and describe scene elements using natural language descriptions rather than predefined category labels. This innovative direction revolutionizes the paradigm of traditional 3D scene analysis, enabling systems to transcend limited predetermined category sets and understand and respond to a rich variety of natural language expressions. By combining visual understanding with the flexibility of language models, open-vocabulary methods can recognize novel objects, understand complex spatial relationships, capture functional attributes of objects, and even infer potential activities within a scene, all without the need to retrain the model for each new concept. This approach not only significantly enhances the system’s ability to handle real-world diversity and uncertainty but also opens new avenues for natural human-machine interaction, allowing users to communicate directly with 3D environments using everyday language, thus providing more intuitive and powerful interaction modes for applications such as augmented reality, robotic navigation, smart environments, and virtual world exploration.

OpenScene~\cite{Peng2023OpenScene} adopts a zero-shot approach, predicting dense features for 3D scene points that are co-embedded with CLIP~\cite{radford2021learning} text and image pixel embeddings in a shared feature space, enabling task-agnostic training and open-vocabulary queries to identify objects, materials, functional affordances, activities, and room types. CLIP-FO3D~\cite{zhang2023clipfo3d} follows a similar approach, modifying CLIP to extract dense pixel features from 3D scenes, which are projected onto point clouds and then used to train a 3D model via distillation to transfer CLIPs knowledge. Semantic abstraction extracts correlation maps from CLIP as abstract object representations to generalize to new semantics, vocabularies, and domains. Open-Fusion~\cite{kashu2023openfusion} combines the SEEM visual-language model with TSDF 3D mapping for real-time open-vocabulary scene creation and querying, leveraging region-based embeddings and confidence maps.

Methods such as PLA~\cite{ding2022language} and RegionPLC~\cite{yang2023regionplc} utilize contrastive learning to integrate descriptive text with 2D and 3D data modalities, associating visual and semantic information. PLA uses 3D-text pairs and contrastive learning to link multi-view images with descriptive text for learning visual-semantic representations, while RegionPLC proposes region-aware contrastive learning by mapping region-level descriptive text from a 2D model to 3D points. OVIR-3D~\cite{lu2023ovir3d} fuses 2D region proposals from off-the-shelf 2D detectors and text-aligned features into 3D instances, achieving efficient open-vocabulary retrieval. CoDA~\cite{cao2023coda} employs 3D geometric priors from annotated base categories and 2D semantic priors from CLIP in its 3D novel object discovery (3D-NOD) strategy. Its discovery-driven cross-modal alignment (DCMA) aligns 3D and image/text features for novel object localization and classification.

Instance-level scene understanding efforts, such as OpenMask3D~\cite{takmaz2023openmask3d}, leverage predicted category-agnostic 3D instance masks and 2D segment-level CLIP embeddings to achieve open-vocabulary 3D instance segmentation. OpenIns3D~\cite{huang2023openins3d} enables open-vocabulary understanding without aligned images, using a “mask-capture-locate” pipeline to predict 3D mask proposals, generate synthetic scene images, and assign categories to masks via a language module. Other work proposes using CLIP features to lay the foundation for 3D feature learning for semantic and instance segmentation. Language grounding with NeRF shows promising results in open-vocabulary scene understanding. Approaches like DFF~\cite{kobayashi2022distilledfeaturefields}, LERF~\cite{lerf2023}, VL-Fields~\cite{tsagkas2023vlfields} and 3D-OVS~\cite{liu20243dovs} distill knowledge from 2D feature extractors like DINO or CLIP into 3D feature fields by minimizing volume rendering feature errors against 2D features, enabling query-based local editing and language grounding in neural implicit representations. LERF~\cite{lerf2023} optimizes dense, scale-conditioned 3D language fields via volume rendering of CLIP embeddings. LangSplat~\cite{qin2023langsplat} and N2F2~\cite{bhalgat2024n2f2} demonstrate efficient open-vocabulary querying and interaction in 3D Gaussian splatting representations by leveraging hierarchical supervision and multi-scale feature fields.

\section{3D Reasoning Segmentation}

\label{sec:define}
\noindent{\bf Problem Definition.}
3D reasoning segmentation task involves generating a 3D segmentation from a given 3D scene point cloud alongside a human-like language instruction. This instruction often demands sophisticated linguistic comprehension, extending beyond mere identification tasks, like 3D referring segmentation task.
For example, rather than processing simple directives like "the red chair," the textual queries might involve intricate descriptions or scenarios, such as "an object usually situated in a living room that can accommodate multiple people sitting together comfortably." which requires in-depth world knowledge and reasoning understanding.

\smallskip{\noindent{\bf Dataset Collection.}}
% Explaining the scarcity of 3D open-vocabulary segmentation data and our approach
Due to the scarcity of standardized datasets for 3D open-vocabulary segmentation, we directly utilize the training and test sets from 3D scene understanding datasets. Specifically, we have collected 3D scans from indoor datasets, such as ScanNetv2, which provides richly annotated RGB-D scans of real-world indoor scenes, encompassing both 2D and 3D data across 1,513 scans. Given the relatively poor performance of open methods in current open-vocabulary segmentation, we adopt the segmentation results generated by the Mask3D model from these 3D scene understanding training sets as the ground truth for comparison.

% Detailing the experimental setup
We conducted experiments on four benchmarks: ScanRefer for single-object visual grounding, Scan2Cap for dense captioning, and both ScanQA and SQA3D for visual question answering. These benchmarks are based on the ScanNet dataset, ensuring consistency in train/validation/test splits, which facilitates joint training and evaluation.

\begin{figure*}[t]
    \centering
    \includegraphics[width=0.975\linewidth]{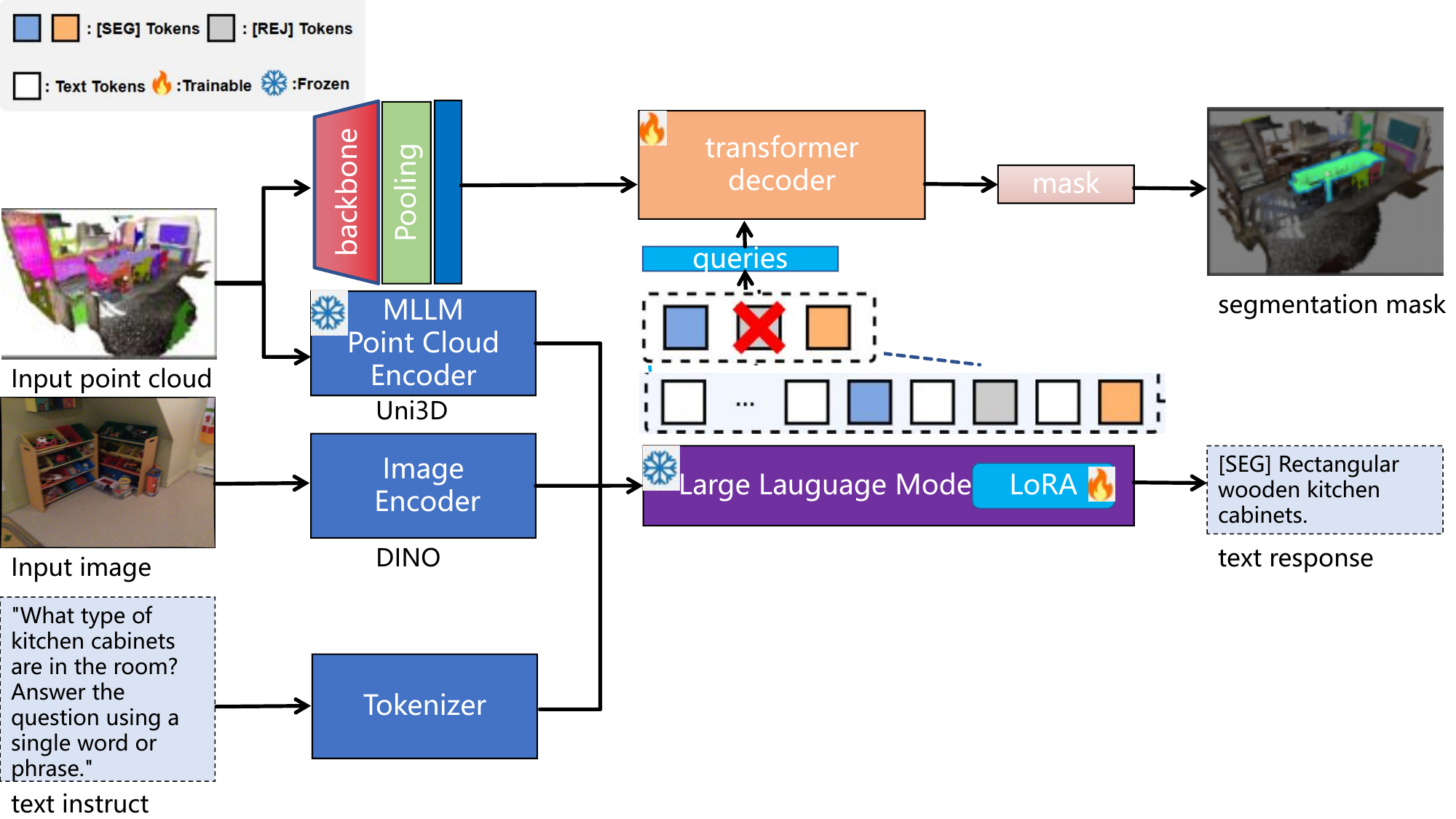}
    \vspace{-1.63cm}
    \vskip 1.5cm
    \caption{\textbf{Overview of our OpenMaskDINO3D framework.} 
    Initially, we utilize a point encoder to extract point features from the input scene, which are simplified by a superpoint pooling layer to reduce complexity. 
    An interactor merges these superpoint features with a learnable query, input into a learnable LLM (trained with LoRA) along with instructions to generate an output containing specifical tokens \textcolor{plotred}{{\texttt{[SEG]}}}.The last-layer embedding for the \textcolor{plotred}{{\texttt{[SEG]}}} token is then decoded into the segmentation mask via the decoder.
    }
\label{fig:arch}
\end{figure*}

\section{OpenMaskDINO3D}
\label{sec:method}

%%%%%%%%%%%%%%%%%%%%%%%%%%%%%%%%%%%%%%%%%%%%%%%%%%%%%%%%%%%%
\subsection{Overview}
The overall architecture of the OpenMaskDINO3D model is depicted in Figure~\ref{fig:arch}. The majority of the OpenMaskDINO3D structure is inherited from spformer, employing the same feature extraction backbone and decoder components. The primary modification lies in enhancing the query selection. Instead of directly utilizing features extracted by the backbone as initial queries, the model uses embeddings from the last layer of a large language model (LLM), processed through a multilayer perceptron (MLP), as the initial queries.

The red component represents the backbone model for extracting supervoxel features, which are obtained via a U-Net structure and aggregated in the green pooling module. The blue component is the feature extraction model, where 3D features of the scene point cloud are extracted using Uni3D, and 2D features are extracted using DINOv2. Text prompts are encoded using the LLM's inherent tokenizer, fused with the 3D and 2D features, and input into the LLM (purple component) for training, ultimately generating the corresponding answer.

Following the approach of LISA, a new token <SEG> is introduced to expand the vocabulary of the original LLM, representing a request for segmentation output. After fusing the 2D features, 3D features, and text features as input to the LLM, the model produces a response text. The output text's tokens include <SEG>, and the embedding $h_{\text{seg}}$ corresponding to the <SEG> token from the last layer is extracted. This embedding is then processed through an MLP projection layer to obtain the required initial query $T_{\text{text}}$. The query $T_{\text{text}}$ associated with <SEG> is used for object segmentation and fed into the orange decoder model. Simultaneously, the supervoxel features extracted by the backbone serve as key-value pairs input to the decoder, forming the "key," "value," and "query" components required for the transformer architecture to perform object segmentation. Ultimately, the model yields the predicted 3D segmentation results.

\subsection{Model Architecture}
\label{sub_sec:alignment} 
As illustrated in Figure~\ref{fig:arch}, our method processes a 3D scene’s point cloud by decomposing it into object proposals using a pre-trained detector. We then employ pre-trained 3D and 2D encoders to derive object-centric representations from point clouds and multi-view images, respectively. These representations are subsequently mapped into the token embedding space of a language model. The text prompts are processed through the tokenizer to generate corresponding token embeddings, which are then combined with the 3D and 2D feature representations. This fused input is fed into a large language model (LLM) for training, enabling the model to generate contextually relevant responses or segmentation outputs based on the integrated multimodal features.

% Scene encoder
\smallskip{\noindent{\bf Object Identifiers.}}
To enable localized understanding of 3D scenes, a set of learnable identifier tokens ${ \texttt{<OBJ}i \texttt{>} }_{i=1}^n$ is introduced into the vocabulary of the language model, designated as object identifiers. These identifiers serve as a bridge between visual representations and linguistic understanding within the system architecture, allowing the model to precisely reference and discuss specific objects in a 3D scene. The learning process for these identifiers occurs concurrently with the training of the entire model, where, through backpropagation, the identifiers gradually acquire semantic associations with their corresponding physical objects. In this manner, object identifiers form the foundation for the model’s object-level reasoning and interaction, facilitating a seamless transition from pixels to semantics and providing richer, more precise expressive capabilities for 3D scene understanding.

\smallskip{\noindent{\bf Uni3D Feature Extraction.}}
We utilizes the Uni3D model to extract 3D features from the point cloud of a given scene, serving as input features for subsequent tasks. To align with the feature fusion process of the LLM model, the extracted 3D features undergo dimensional pruning to match the input feature dimensions of the LLM model. This embedding process handles the point cloud $P_i$ of each object, producing an output feature $Z_i^p$ for each object.

\smallskip{\noindent{\bf DINOv2 Feature Extraction.}}
We employs a method that projects the point cloud of each object onto multi-view images, creating a series of 2D masks. Using the pre-trained DINOv2 model, local features are extracted and aggregated from the masked regions of the multi-view images for each object, taking into account both the mask regions and multi-view information. The 2D encoder processes the multi-view images and their corresponding projected masks derived from the point cloud $P_i$ of each object, generating visual features $Z_i^v$ for each object.

\smallskip{\noindent{\bf Feature Fusion.}}
To effectively integrate object information from different modalities, we proposes a feature fusion mechanism based on interleaving. This mechanism uses object identifiers as the backbone to systematically organize multimodal features into a sequential representation for processing by the language model. Specifically, given the object identifier embedding $ O_i $, 3D point cloud features $ F_i^p $, 2D visual features $ F_i^v $, and positional encoding $ L_i $, an interleaved sequence is constructed as follows:
$$S_i = [O_i, F_i^p, F_i^v, L_i] \quad $$
Here, the object identifier embedding $ O_i $ appears first, followed by the various feature representations. This interleaved arrangement enables the language model to effectively associate the identifier with its corresponding visual and spatial features, establishing semantic connections between them while preserving their individual characteristics.

\subsection{Training OpenMaskDINO3D} 
\label{sub_sec:training} 
The loss function of OpenMaskDINO3D consists of two fundamental components: the LLM loss $ L_{\text{llm}} $ and the segmentation mask loss $ L_{\text{mask}} $. The overall loss is expressed as:
$$L = L_{\text{llm}} + L_{\text{mask}} \quad $$
The LLM loss $ L_{\text{llm}} $ captures the semantic consistency of language generation through an autoregressive cross-entropy loss. Since all tasks in this paper have been unified into a consistent user-assistant interaction format, this loss is optimized for the negative log-likelihood of the target response sequence $ s_{\text{res}} $:
$$L_{\text{llm}} = -\sum_{i=1}^k \log P(s_i^{\text{res}} | s_{[1,\ldots,i-1]}^{\text{res}}, s_{\text{prefix}}) \quad $$
Here, $ k $ denotes the number of tokens in the response sequence, $ s_{[1,\ldots,i-1]}^{\text{res}} $ represents the sequence of the first $ i-1 $ tokens in the response, and $ s_{\text{prefix}} $ is the input prefix sequence containing system messages and user instructions. The trainable parameters include the vision-language projector, the token embeddings of object identifiers, and the language model parameters.
Meanwhile, the mask loss $ L_{\text{mask}} $ aims to encourage the model to generate high-quality segmentation masks, computed using binary cross-entropy (BCE) loss and DICE loss for region-level and object-level supervoxel segmentation:
$$L_{\text{mask}}^* = \text{BCE}(M^*, \hat{M}^*) + \text{DICE}(M^*, \hat{M}^*), \quad * \in [\text{seg}] \quad $$
Here, $ M^* $ represents the ground-truth segmentation mask, and $ \hat{M}^* $ denotes the predicted mask. For object-level masks $ M_{\text{seg}} $, the mask of the target object is used.

\section{Experiments}

%%%%%%%%%%%%%%%%%%%%%%%%%%%%%%%%%%%%%%%%%%%%%%%%%%%%%%%%%%%%

\begin{table*}[t]
    \footnotesize
    \centering
    \begin{tabular}{l|ccc}
    \hline
    \multirow{2}{*}{Method} & \multicolumn{3}{c}{ScanNet} \\
    \cline{2-4}
     & 0.25 & 0.50 & mIoU \\
    OpenScene [31] & 4.22 & 0.97 & 5.03 \\
    OpenMask3D [39] & 5.70 & 3.25 & 7.14 \\
    3D-STMN [45] & 25.43 & 17.78 & 18.23 \\
    Reason3D & 43.21 & 32.10 & 31.20 \\
    OpenMaskDINO3D & 54.21 & 39.14 & 39.81 \\
    \end{tabular}
    \caption{Performance comparison on Segmentation Tasks.}
    \label{tab:segperformance}
\end{table*}

\begin{table*}[t]
\footnotesize
\centering
\begin{tabular}{l|l|cc|cc|cc|cc}
\hline
 & Method & \multicolumn{2}{c|}{ScanRefer} & \multicolumn{2}{c|}{Scan2Cap} & \multicolumn{2}{c|}{ScanQA} & \multicolumn{2}{c}{SQA3D} \\
\hline
 & & Acc@0.25 & Acc@0.5 & C@0.5 & B-4@0.5 & C & B-4 & EM & EM-R \\
\hline
Expert Models & ScanRefer & 37.3 & 24.3 & - & - & - & - & - & - \\
 & ScanQA & - & - & - & - & 64.9 & 10.1 & - & - \\
 & 3DJCG & 49.6 & 37.3 & 49.5 & 31.0 & - & - & - & - \\
 & 3D-VLP & 51.4 & 39.5 & 54.9 & 32.3 & 67.0 & 11.1 & - & - \\
 & M3DRef-CLIP & 51.9 & 44.7 & - & - & - & - & - & - \\
 & 3D-VisTA & 50.6 & 45.5 & 66.9 & 34.0 & 72.9 & 13.1 & 48.5 & - \\
 & ConcreteNet & 50.6 & 46.5 & - & - & - & - & - & - \\
 & Vote2Cap-DETR++ & - & - & 67.6 & 37.1 & - & - & - & - \\
\hline
LLM-based Models & LAMM & - & 3.4 & - & - & 42.4 & 5.8 & - & - \\
 & Chat-3D & - & - & - & - & 53.2 & 6.4 & - & - \\
 & 3D-LLM & 30.3 & - & - & - & 69.4 & 12.0 & - & - \\
 & LL3DA & - & - & 65.2 & 36.8 & 76.8 & 13.5 & - & - \\
 & LEO & - & - & 68.4 & 36.9 & 80.0 & 11.5 & - & 53.7 \\
 & Scene-LLM & - & - & - & - & 80.0 & 12.0 & 54.2 & - \\
 & OpenMaskDINO3D & 42.3 & 38.3 & 70.2 & 31.0 & 75.6 & 10.7 & - & 44.1 \\
\hline
\end{tabular}
\vspace{-2mm}
\caption{Performance comparison on QA.}
\label{tab:qacomparison}
\end{table*}

\subsection{Experimental Setting}

\noindent{\bf Datasets.} 
We conducted experiments on four benchmarks: ScanRefer for single-object visual grounding, Scan2Cap [12] for dense captioning, and both ScanQA and SQA3D for visual question answering. These benchmarks are based on the ScanNet dataset , which comprises richly annotated RGB-D scans of real-world indoor scenes, including both 2D and 3D data across 1,513 scans. All benchmarks adhere to the same train/validation/test splits, facilitating joint training and evaluation.

\smallskip{\noindent{\bf Evaluation Metrics.}} 
For the tasks of 3D expressive referring segmentation and 3D reasoning segmentation, the primary evaluation metrics are the Mean Intersection over Union (mIoU) and Accuracy at $k$ Intersection over Union (Acc@kIoU). The mIoU quantifies the average overlap between the predicted and ground-truth 3D volumes, providing a measure of segmentation quality. The Acc@kIoU metric evaluates the proportion of descriptions for which the predicted mask overlaps with the ground truth at an IoU greater than $k$, with thresholds set at $k = 0.25$ and $k = 0.5$. These thresholds assess the model’s performance across varying levels of precision.

To evaluate scene understanding, we adhere to established metrics for relevant benchmarks. For ScanRefer, we measure thresholded accuracy with Acc@0.25 and Acc@0.5, where predictions are considered positive if their Intersection over Union (IoU) with the ground truth exceeds the thresholds of 0.25 and 0.5, respectively. For Scan2Cap, we utilize CIDEr@0.5 and BLEU-4@0.5 (abbreviated as C@0.5 and B-4@0.5), combining image captioning metrics with IoU scores between predicted and target bounding boxes. For ScanQA, the metrics CIDEr and BLEU-4 (abbreviated as C and B-4) are used. For SQA3D, evaluation is based on exact match accuracy (EM) and its refined version, EM-R, as proposed by LEO.
%

%%%%%%%%%%%%%%%%%%%%%%%%%%%%%%%%%%%%%%%%%%%%%%%%%%%%%%%%%%%%

\subsection{3D Scene Understanding Results}

OpenMaskDINO3D demonstrates strong performance across various 3D understanding tasks. In the ScanRefer visual grounding task, the model achieves accuracies of 42.3\% (Acc@0.25) and 38.3\% (Acc@0.5). While these results are lower than those of the state-of-the-art Chat-Scene model, they significantly outperform certain earlier LLM models, such as 3D-LLM. In the Scan2Cap dense captioning task, OpenMaskDINO3D attains scores of 70.2 (C@0.5) and 31.0 (B-4@0.5), indicating its robust capability in generating descriptive text with spatial localization. For the ScanQA visual question answering task, the model achieves scores of 75.6 (C) and 10.7 (B-4), reflecting moderate performance. On the SQA3D dataset, OpenMaskDINO3D obtains an EM-R score of 44.1\%, demonstrating competitive performance in complex scene understanding question answering. Overall, OpenMaskDINO3D exhibits comprehensive capabilities across all evaluated tasks. Although it does not surpass the state-of-the-art Chat-Scene in most metrics, it consistently outperforms many specialized models and earlier LLM models, highlighting its potential as a versatile 3D scene understanding system. Detailed results are presented in Table ~\ref{tab:qacomparison}.

\subsection{3D Reasoning Segmentation Results}
Based on the table data, OpenMaskDINO3D demonstrates outstanding performance in 3D segmentation tasks, as detailed in Table ~\ref{tab:segperformance}. On the ScanNet dataset, OpenMaskDINO3D achieves strong results across all evaluation metrics, including Acc@0.25 (54.21\%), Acc@0.50 (39.14\%), and mIoU (39.81\%). These values significantly surpass those of other models, including OpenScene, OpenMask3D, 3D-STMN, and Reason3D. Compared to the second-best model, Reason3D, OpenMaskDINO3D improves by approximately 11 percentage points in Acc@0.25 (54.21\% vs. 43.21\%), 7 percentage points in Acc@0.50 (39.14\% vs. 32.10\%), and 8.6 percentage points in mIoU (39.81\% vs. 31.20\%). The improvements over earlier models such as OpenScene and OpenMask3D are even more pronounced, reflecting the rapid advancements in 3D segmentation technology. For instance, compared to OpenMask3D, OpenMaskDINO3D’s mIoU value increases by over 32 percentage points (39.81\% vs. 7.14\%).

\subsection{Ablation Study}
\smallskip{
\noindent{\bf Effectiveness of 2D Features.}}
Regarding the impact of 2D features on question-answering performance, the table data indicates that 2D feature extraction plays a significant role in 3D scene understanding. By introducing additional visual cues, the algorithm achieves substantial improvements across metrics such as accuracy, capture rate, and CIDEr. Specifically, the incorporation of 2D features boosts accuracy from 27.8\% to 36.8\%, increases the capture rate from 18.4\% to 27.5\%, and enhances the CIDEr score from 23.3 to 27.0, highlighting the critical role of cross-modal feature fusion in enhancing algorithmic performance. Detailed results are presented in Table ~\ref{tab:2d_features}(a).

According to Table ~\ref{tab:2d_features}(b), the introduction of 2D features has a significant positive impact on segmentation performance. Performance metrics show notable improvements across different thresholds: at the 0.25 threshold, performance increases from 51.4\% to 55.1\%; at the 0.5 threshold, it rises from 35.5\% to 40.2\%; and the mean Intersection over Union (mIoU) improves from 36.7\% to 40.3\%. These results demonstrate that incorporating 2D image features effectively enhances the precision and robustness of the segmentation algorithm, underscoring the critical role of cross-modal feature fusion in improving segmentation performance.

\begin{table*}[t]
\vspace{-.2em}
\centering
%#################################################
\hspace{-0.5em}
\subfloat[
{2D Features for Question Answering}.
\label{tab:image_token_qa}
]{
\centering
\begin{minipage}{0.45\linewidth}{
\begin{center}
\tablestyle{2pt}{1.00}
\scriptsize
\begin{tabular}{ccccc} \toprule
Image Token & ScanRefer (Acc@0.5) & Scan2Cap (C@0.5) & ScanQA (CIDEr) & SQA3D (EM) \\
\midrule
\xmark & 27.8 & 18.4 & 23.3 & 44.9 \\
\cmark & 36.8 & 27.5 & 27.0 & 42.6 \\
\bottomrule
\end{tabular}
\end{center}}\end{minipage}
}
\hspace{1.2em}
%#################################################
\subfloat[
{2D Features for Segmentation}.
\label{tab:image_token_seg}
]{
\begin{minipage}{0.35\linewidth}{\begin{center}
\tablestyle{2pt}{1.00}
\scriptsize
\begin{tabular}{cccc} \toprule
Image Token & Acc@0.25 & Acc@0.50 & mIoU \\
\midrule
\xmark & 51.4 & 35.5 & 36.7 \\
\cmark & 55.1 & 40.2 & 40.3 \\
\bottomrule
\end{tabular}
\end{center}}\end{minipage}
}
%#################################################
\caption{\textbf{Impact of 2D features} on question-answering and segmentation tasks, evaluated on multiple datasets.} \vspace{-0.5em}
\label{tab:2d_features}
\end{table*}

\begin{table*}[t]
\vspace{-.2em}
\centering
%#################################################
\hspace{-0.5em}
\subfloat[
{Object Identifiers for Question Answering}.
\label{tab:object_token_qa}
]{
\centering
\begin{minipage}{0.45\linewidth}{
\begin{center}
\tablestyle{2pt}{1.00}
\scriptsize
\begin{tabular}{ccccc} \toprule
Object Token & ScanRefer (Acc@0.5) & Scan2Cap (C@0.5) & ScanQA (CIDEr) & SQA3D (EM) \\
\midrule
\xmark & 34.9 & 26.7 & 26.5 & 51.3 \\
\cmark & 36.8 & 27.5 & 27.0 & 42.6 \\
\bottomrule
\end{tabular}
\end{center}}\end{minipage}
}
\hspace{1.2em}
%#################################################
\subfloat[
{Object Identifiers for Segmentation}.
\label{tab:object_token_seg}
]{
\begin{minipage}{0.35\linewidth}{\begin{center}
\tablestyle{2pt}{1.00}
\scriptsize
\begin{tabular}{cccc} \toprule
Object Token & Acc@0.25 & Acc@0.50 & mIoU \\
\midrule
\xmark & 51.4 & 36.1 & 37.0 \\
\cmark & 55.1 & 40.2 & 40.3 \\
\bottomrule
\end{tabular}
\end{center}}\end{minipage}
}
%#################################################
\caption{\textbf{Impact of object identifiers} on question-answering and segmentation tasks, evaluated on multiple datasets.} \vspace{-0.5em}
\label{tab:object_identifiers}
\end{table*}

\smallskip{
\noindent{\bf Effectiveness of Object Identifiers.}}
According to the table data, the introduction of object identifiers has a positive impact on question-answering tasks. Across the ScanRefer, Scan2Cap, and ScanQA metrics, the inclusion of object identifiers results in slight improvements: accuracy (Acc@0.5) increases from 34.9\% to 36.8\%, capture rate (C@0.5) rises from 26.7\% to 27.5\%, and the CIDEr score improves from 26.5 to 27.0. However, the exact match (EM) metric exhibits a significant decline, dropping from 51.3\% to 42.6\%. This indicates that while object identifiers enhance performance in multimodal scene understanding tasks, further optimization is needed for exact match performance. Detailed results are presented in Table~\ref{tab:object_identifiers}(a).

Object identifiers demonstrate a significant positive impact on segmentation performance. As shown in Table~\ref{tab:object_identifiers}(b), notable improvements are observed across all evaluation metrics: with the use of object identifiers, performance at the 0.25 threshold increases from 51.4\% to 55.1\% (a 3.7 percentage point gain); at the 0.5 threshold, performance rises from 36.1\% to 40.2\% (a 4.1 percentage point gain); and in terms of mean Intersection over Union (mIoU), performance improves from 37.0\% to 40.3\% (a 3.3 percentage point gain).

\section{Conclusion}
This study introduces OpenMaskDINO3D, a point cloud segmentation framework guided by Large Language Models (LLMs) to improve 3D scene understanding and generate precise segmentation masks. We propose a novel [SEG] token for high-accuracy 3D mask generation and object identifiers (<OBJk>) for unified object referencing, addressing semantic ambiguity. A superpoint pooling layer and hierarchical mask decoder tackle large-scale point cloud challenges, enabling coarse-to-fine localization. By integrating Uni3D and DINOv2 features with datasets like ScanRefer, Scan2Cap, ScanQA, and SQA3D, the model achieves robust training. Experiments on ScanNet show superior performance in 3D segmentation (Acc@0.25: 54.21\%, Acc@0.50: 39.14\%, mIoU: 39.81\%), outperforming many specialized models, with ablation studies confirming the value of 2D features and object identifiers.

{
    \small
    \bibliographystyle{ieeenat_fullname}
    \bibliography{ref}

\begin{thebibliography}{16}
\providecommand{\natexlab}[1]{#1}
\providecommand{\url}[1]{\texttt{#1}}
\expandafter\ifx\csname urlstyle\endcsname\relax
  \providecommand{\doi}[1]{doi: #1}\else
  \providecommand{\doi}{doi: \begingroup \urlstyle{rm}\Url}\fi

\bibitem[Bhalgat et~al.(2024)Bhalgat, Laina, Henriques, Zisserman, and Vedaldi]{bhalgat2024n2f2}
Yash Bhalgat, Iro Laina, João~F. Henriques, Andrew Zisserman, and Andrea Vedaldi.
\newblock N2f2: Hierarchical scene understanding with nested neural feature fields, 2024.

\bibitem[Cao et~al.(2023)Cao, Yihan, Xu, and Xu]{cao2023coda}
Yang Cao, Zeng Yihan, Hang Xu, and Dan Xu.
\newblock Coda: Collaborative novel box discovery and cross-modal alignment for open-vocabulary 3d object detection.
\newblock \emph{NIPS}, 36, 2023.

\bibitem[Ding et~al.(2023)Ding, Yang, Xue, Zhang, Bai, and Qi]{ding2022language}
Runyu Ding, Jihan Yang, Chuhui Xue, Wenqing Zhang, Song Bai, and Xiaojuan Qi.
\newblock Pla: Language-driven open-vocabulary 3d scene understanding.
\newblock In \emph{CVPR}, 2023.

\bibitem[Huang et~al.(2023)Huang, Wu, Chen, Zhao, Zhu, and Lasenby]{huang2023openins3d}
Zhening Huang, Xiaoyang Wu, Xi Chen, Hengshuang Zhao, Lei Zhu, and Joan Lasenby.
\newblock Openins3d: Snap and lookup for 3d open-vocabulary instance segmentation.
\newblock \emph{arXiv preprint}, 2023.

\bibitem[Kerr et~al.(2023)Kerr, Kim, Goldberg, Kanazawa, and Tancik]{lerf2023}
Justin* Kerr, Chung~Min* Kim, Ken Goldberg, Angjoo Kanazawa, and Matthew Tancik.
\newblock Lerf: Language embedded radiance fields.
\newblock In \emph{ICCV}, 2023.

\bibitem[Kobayashi et~al.(2022)Kobayashi, Matsumoto, and Sitzmann]{kobayashi2022distilledfeaturefields}
Sosuke Kobayashi, Eiichi Matsumoto, and Vincent Sitzmann.
\newblock Decomposing nerf for editing via feature field distillation.
\newblock \emph{NIPS}, 35:\penalty0 23311--23330, 2022.

\bibitem[Liu et~al.(2024)Liu, Zhan, Zhang, Xu, Yu, El~Saddik, Theobalt, Xing, and Lu]{liu20243dovs}
Kunhao Liu, Fangneng Zhan, Jiahui Zhang, Muyu Xu, Yingchen Yu, Abdulmotaleb El~Saddik, Christian Theobalt, Eric Xing, and Shijian Lu.
\newblock Weakly supervised 3d open-vocabulary segmentation.
\newblock \emph{NIPS}, 36, 2024.

\bibitem[Lu et~al.(2023)Lu, Chang, Jing, Boularias, and Bekris]{lu2023ovir3d}
Shiyang Lu, Haonan Chang, Eric~Pu Jing, Abdeslam Boularias, and Kostas Bekris.
\newblock Ovir-3d: Open-vocabulary 3d instance retrieval without training on 3d data.
\newblock In \emph{Conference on Robot Learning}, pages 1610--1620. PMLR, 2023.

\bibitem[Peng et~al.(2023)Peng, Genova, Jiang, Tagliasacchi, Pollefeys, and Funkhouser]{Peng2023OpenScene}
Songyou Peng, Kyle Genova, Chiyu~"Max" Jiang, Andrea Tagliasacchi, Marc Pollefeys, and Thomas Funkhouser.
\newblock Openscene: 3d scene understanding with open vocabularies.
\newblock In \emph{CVPR}, 2023.

\bibitem[Qin et~al.(2023)Qin, Li, Zhou, Wang, and Pfister]{qin2023langsplat}
Minghan Qin, Wanhua Li, Jiawei Zhou, Haoqian Wang, and Hanspeter Pfister.
\newblock Langsplat: 3d language gaussian splatting.
\newblock \emph{arXiv preprint arXiv:2312.16084}, 2023.

\bibitem[Radford et~al.(2021)Radford, Kim, Hallacy, Ramesh, Goh, Agarwal, Sastry, Askell, Mishkin, Clark, et~al.]{radford2021learning}
Alec Radford, Jong~Wook Kim, Chris Hallacy, Aditya Ramesh, Gabriel Goh, Sandhini Agarwal, Girish Sastry, Amanda Askell, Pamela Mishkin, Jack Clark, et~al.
\newblock Learning transferable visual models from natural language supervision.
\newblock In \emph{International conference on machine learning}, pages 8748--8763. PmLR, 2021.

\bibitem[Takmaz et~al.(2023)Takmaz, Fedele, Sumner, Pollefeys, Tombari, and Engelmann]{takmaz2023openmask3d}
Ay{\c{c}}a Takmaz, Elisabetta Fedele, Robert~W. Sumner, Marc Pollefeys, Federico Tombari, and Francis Engelmann.
\newblock {OpenMask3D: Open-Vocabulary 3D Instance Segmentation}.
\newblock In \emph{NeurIPS}, 2023.

\bibitem[Tsagkas et~al.(2023)Tsagkas, Mac~Aodha, and Lu]{tsagkas2023vlfields}
Nikolaos Tsagkas, Oisin Mac~Aodha, and Chris~Xiaoxuan Lu.
\newblock Vl-fields: Towards language-grounded neural implicit spatial representations.
\newblock \emph{arXiv preprint arXiv:2305.12427}, 2023.

\bibitem[Yamazaki et~al.(2023)Yamazaki, Hanyu, Vo, Pham, Tran, Doretto, Nguyen, and Le]{kashu2023openfusion}
Kashu Yamazaki, Taisei Hanyu, Khoa Vo, Thang Pham, Minh Tran, Gianfranco Doretto, Anh Nguyen, and Ngan Le.
\newblock Open-fusion: Real-time open-vocabulary 3d mapping and queryable scene representation.
\newblock \emph{arXiv preprint arXiv:2310.03923}, 2023.

\bibitem[Yang et~al.(2023)Yang, Ding, Wang, and Qi]{yang2023regionplc}
Jihan Yang, Runyu Ding, Zhe Wang, and Xiaojuan Qi.
\newblock Regionplc: Regional point-language contrastive learning for open-world 3d scene understanding.
\newblock \emph{arXiv preprint arXiv:2304.00962}, 2023.

\bibitem[Zhang et~al.(2023)Zhang, Dong, and Ma]{zhang2023clipfo3d}
Junbo Zhang, Runpei Dong, and Kaisheng Ma.
\newblock Clip-fo3d: Learning free open-world 3d scene representations from 2d dense clip.
\newblock In \emph{ICCV}, pages 2048--2059, 2023.

\end{thebibliography}
}

% \maketitlesupplementary
% \renewcommand\thesection{\Alph{section}}
% \setcounter{section}{0}
% \input{sec/X_suppl}

\end{document}